\documentclass[10pt,twocolumn,letterpaper]{article}
\pdfoutput=1 
\usepackage{iccv}
\usepackage{multirow}
\usepackage{caption}
\usepackage{subcaption}
\expandafter\def\csname ver@subfig.sty\endcsname{}
\usepackage{graphicx,subfig}

\usepackage{color}
\definecolor{green}{rgb}{0, 0.5, 0}
\definecolor{orange}{rgb}{0.8, 0.6, 0.2}
\definecolor{orange2}{rgb}{1.0, 0.6, 0.2}
\definecolor{red}{rgb}{1.0, 0.0, 0.0}
\definecolor{teal}{rgb}{0.0, 0.4, 0.4}
\definecolor{purple}{rgb}{0.65,0,0.65}
\definecolor{saffron}{rgb}{0.95,0.75,0.2}
\definecolor{turquoise}{rgb}{0.0,0.5,0.5}
\definecolor{black}{rgb}{0.0, 0.0, 0.0}
\definecolor{gray}{rgb}{0.5, 0.5, 0.5}
\definecolor{softblue}{RGB}{0,122,204}

\definecolor{iccvblue}{rgb}{0.21,0.49,0.74}
\usepackage[pagebackref,breaklinks,colorlinks,allcolors=iccvblue]{hyperref}

\title{LAMIC: Layout-Aware Multi-Image Composition via Scalability of Multimodal Diffusion Transformer}

\renewcommand{\thefootnote}{\fnsymbol{footnote}}
\author{
Yuzhuo Chen$^{1}$ \quad Zehua Ma$^{1}$ \quad Jianhua Wang$^{2}$ \quad Kai Kang$^{3}$ 
\quad Shunyu Yao$^{2}$\thanks{Project Leader}
\quad Weiming Zhang$^{1}$ \\
$^{1}$University of Science and Technology of China \\
$^{2}$Onestory Team \\
$^{3}$East China Normal University
\\
\url{http://github.com/Suchenl/LAMIC}
}

\begin{document}

\twocolumn[{
\renewcommand\twocolumn[1][]{#1}
\maketitle
\vspace{-1.0cm} 
\begin{center}
    \captionsetup{type=figure}
    \includegraphics[width=0.95\textwidth]{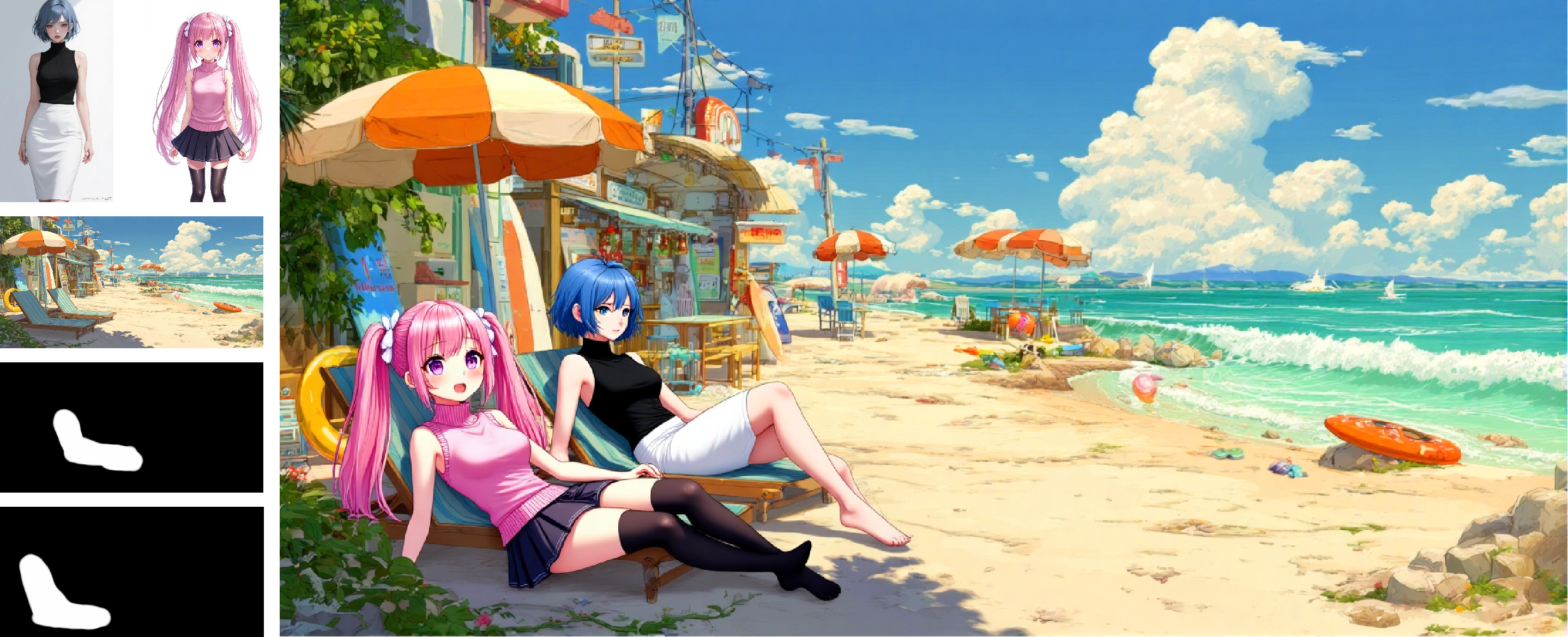}
    \captionof{figure}{An example of layout-aware multi-image composition generated by our proposed model, LAMIC.}
\end{center}
\vspace{-0.01cm} 
}]

\def\thefootnote{†}\footnotetext{Project Leader}

\begin{abstract}
In controllable image synthesis, generating coherent and consistent images from multiple references with spatial layout awareness remains an open challenge.
We present \textbf{LAMIC}, a \textbf{L}ayout-\textbf{A}ware \textbf{M}ulti-\textbf{I}mage \textbf{C}omposition framework that, for the first time, extends single-reference diffusion models to multi-reference scenarios in a training-free manner. Built upon the MMDiT model, LAMIC introduces two plug-and-play attention mechanisms: 1) Group Isolation Attention (GIA) to enhance entity disentanglement; and 2) Region-Modulated Attention (RMA) to enable layout-aware generation. To comprehensively evaluate model capabilities, we further introduce three metrics: 1) Inclusion Ratio (IN-R) and Fill Ratio (FI-R) for assessing layout control; and 2) Background Similarity (BG-S) for measuring background consistency. 
Extensive experiments show that LAMIC achieves state-of-the-art performance across most major metrics: it consistently outperforms existing multi-reference baselines in ID-S, BG-S, IN-R and AVG scores across all settings, and achieves the best DPG in complex composition tasks. 
These results demonstrate LAMIC's superior abilities in identity keeping, background preservation, layout control, and prompt-following, all achieved without any training or fine-tuning, showcasing strong zero-shot generalization ability. 
By inheriting the strengths of advanced single-reference models and enabling seamless extension to multi-image scenarios, LAMIC establishes a new training-free paradigm for controllable multi-image composition. As foundation models continue to evolve, LAMIC's performance is expected to scale accordingly. Our implementation is available at: {\normalfont\href{https://github.com/Suchenl/LAMIC}{\textcolor{softblue}{\textit{\underline{https://github.com/Suchenl/LAMIC}}}}}.

\end{abstract}


\section{Introduction}

Creating consistent and controllable visual content is a core challenge in digital filmmaking, storyboarding, and narrative illustration~\cite{zhou2024storydiffusion}. In these domains, artists often need to construct scenes that involve multiple entities—such as characters and environments—while maintaining visual and stylistic consistency across varying perspectives and story beats. With the rapid progress of diffusion-based generative models, there is growing interest in leveraging such models to automate and accelerate the generation of entity-consistent, layout-controllable images guided by both textual and visual inputs~\cite{xiao2025captain_cinema}.

Recent advances in image generation have introduced impressive capabilities in text-to-image (T2I) and image-to-image (I2I) synthesis~\cite{rombach2022latent, brooks2023instructpix2pix}. More recently, multimodal text-and-image-to-image (T\&I2I) models, especially FLUX.1-Kontext released on May 29, 2025~\cite{DMs-FluxKontext}, have demonstrated the great potential of combining textual descriptions with visual references to generate semantically grounded and identity-consistent images. However, as a single-reference model, it still remains significantly limited in handling multiple reference images.
To enable multi-image reference generation, some methods have introduced trainable extension modules into the fundamental T2I models~\cite{chen2025xverse}, while others have retrained a slightly modified T2I architecture~\cite{xiao2024omnigen,wu2025omnigen2}. However, these training-based methods face challenges in generalization performance when combining more images, as large-scale datasets with multiple image references are difficult to collect~\cite{chen2025xverse}. 

In addition, many of these methods lack spatial layout capabilities, which limits their application in real scenarios. These limitations become particularly problematic in creative production workflows. For instance, in storyboard generation for films or animations, it is crucial to consistently generate the same multiple characters, objects and scenes which conform to explicit layout plans-such as character positioning, scene framing, or camera perspective determined by directors or artists.
Previous studies have investigated layout control in T2I systems, which can be mainly divided into training-based ~\cite{DMs-Easycontrol_2025, DMs-Ominicontrol2_2025} and training-free ~\cite{yang2024mastering-RPG, chen2024region-RAG}. However, the former requires the introduction of additional modules for specific tasks or fine-tuning with LoRA to force the generated image to conform to the layout input, which is still constrained by the dataset. The latter is mainly achieved by manipulating the area where the prompt word is injected or using the model's local predicted noise to replace the corresponding area in the global noise predicted by the model. However, such methods are prone to cross-image interference and semantic leakage, especially in the case of similar appearance of entities (for example, multiple humans or animals), resulting in reduced subject consistency.

To address the above limitations, we propose \textbf{LAMIC (Layout-Aware Multi-Image Composition)}, a training-free framework built upon a pretrained single-image reference Multimodal Diffusion Transformer (MMDiT)~\cite{DMs-SD3} model. Leveraging our proposed attention mechanism and the strong scalability of the MMDiT architecture, LAMIC enables users to incorporate an arbitrary number of reference images along with region-level layout priors (e.g., masks or bounding boxes). Our framework achieves superior overall performance compared to prior approaches in multi-image composition, particularly excelling in tasks involving fine-grained layout control. The main contributions of this paper are as follows:
\begin{itemize}
    \item We propose \textbf{LAMIC}, a novel framework for high-quality \textbf{layout-aware multi-image composition}, supporting flexible spatial layout control and seamless multi-reference integration.
    
    \item LAMIC is the first to \textbf{extend single-reference diffusion models to multi-reference scenarios} in a \textbf{training-free} manner. It inherits the consistency-preserving editing capabilities of the underlying model, while circumventing the generalization issues caused by the scarcity of large-scale multi-reference training datasets.
    

    \item To reduce semantic entanglement across entities, we introduce \textbf{Group Isolation Attention (GIA)}, which enforces localized attention within aligned visual-textual-spatial (VTS) triplets.
    
    \item Building on GIA, we further propose \textbf{Region-Modulated Attention (RMA)}, which defers inter-region fusion and cross-entity interaction (CEI) instruction injection to enhance layout controllability and prevent early-stage semantic leakage.
\end{itemize}





\section{Related Work}
\paragraph{Reference-Guided Image Generation.}
Recent advancements in multimodal image generation enable synthesizing images guided by both textual and visual references. UniAdapter~\cite{wang2023uniadapter} introduced early adapter-based methods for reference adaptation, though limited to single references without spatial disentanglement. EasyRef~\cite{zong2024easyref} proposed leveraging multiple reference images but relies on complex multimodal large language models (MLLMs), hindering practical usability. FLUX.1-Kontext~\cite{DMs-FluxKontext}, built on MMDiT architecture, demonstrates significant improvements in identity consistency using a single reference image. However, these methods remain inadequate for handling multiple image references effectively.

\begin{figure*}[t!]
    \centering
    \includegraphics[width=\linewidth]{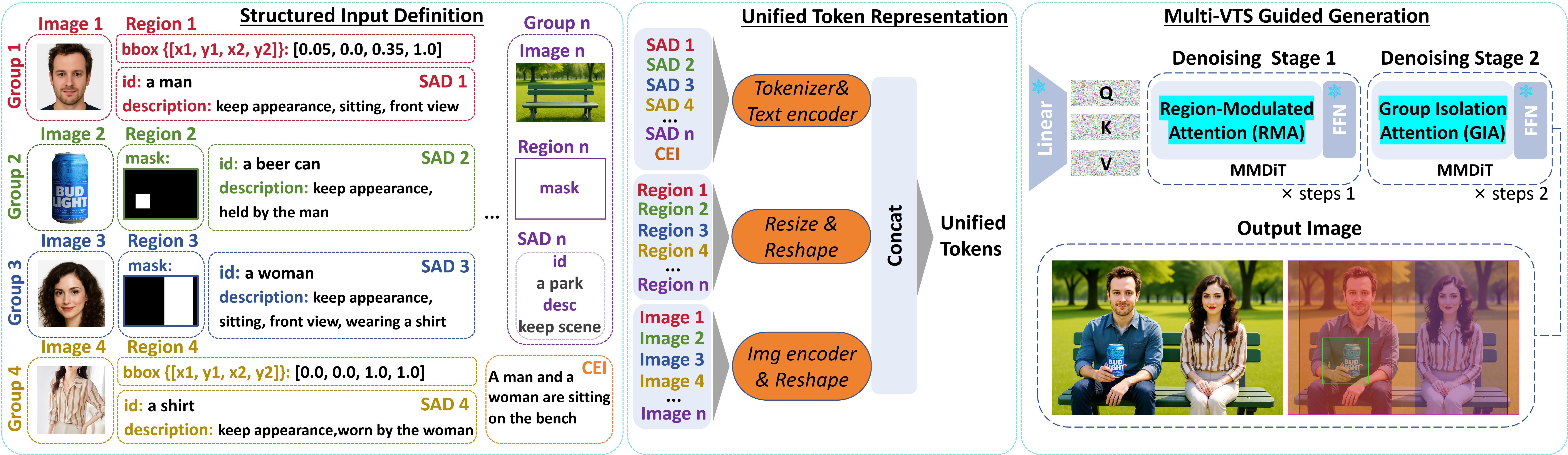}
    \caption{Framework of our proposed LAMIC. We illustrate the layout-aware multi-image composition process with 5 reference groups (n=5) provided as input.}
    \label{fig:framework}
\end{figure*}

\paragraph{Layout-Aware Generation.}
Spatial layout control has been explored via supervised segmentation maps, bounding-box conditioning~\cite{chen2024region-RAG}, and region-aware prompts. However, most approaches rely on either fine-tuning~\cite{ruiz2023dreambooth}, prompt heuristics~\cite{yang2024mastering-RPG}, training-time supervision~\cite{he2024eligen} or repeated inference ~\cite{chen2024region-RAG}, making them less flexible for open-domain generation. In contrast, our method avoids parameter tuning, extra inference, and complex prompt engineering, offering a more practical solution for open-domain scenarios.

\paragraph{Multi-Image Composition.}
Effective compositional generation involves integrating multiple visual references into coherent images. MS-Diffusion~\cite{wang2025msdiffusion} pioneered multi-modal inference with layout control but exhibits limitations in identity preservation and spatial accuracy. Methods like OmniGen~\cite{xiao2024omnigen} and OmniGen2~\cite{wu2025omnigen2} enhance identity disentanglement but generally require extensive retraining, restricting scalability. UNO~\cite{wu2025less_UNO} and DreamO~\cite{mou2025dreamo} provide consistent cross-reference synthesis but lack explicit layout control. Although XVerse~\cite{chen2025xverse} achieves fine-grained identity control, like most training-based methods, it relies on large-scale multi-reference datasets that are difficult to collect, leading to generalization limitations in practical scenarios.

\section{Method} 
\subsection{Preliminaries and Key Insights}
\paragraph{Multimodal Diffusion Transformer.}
MMDiT~\cite{DMs-SD3}, introduced in Stable Diffusion 3, extends DiT~\cite{DMs-DiT} by concatenating text and latent image tokens for multimodal conditioning within the LDM framework~\cite{DMs-LDMs}. This design has been adopted in models such as FLUX.1~\cite{DMs-Flux1} and FLUX.1-Kontext~\cite{DMs-FluxKontext}, with the latter demonstrating strong identity preservation in single-reference generation. Notably, both Kontext and recent control frameworks (e.g., EasyControl~\cite{DMs-Easycontrol_2025}, OmniControl2~\cite{DMs-Ominicontrol2_2025}) introduce external control signals—like reference images—via token concatenation. This design paradigm reveals a \textbf{key insight}: it is possible to introduce multiple-reference images into a unified representation space using only a pretrained single-reference network.

\paragraph{Attention Mechanism.}
In MMDiT, both self-attention and cross-attention layers are applied at each denoising step to model complex dependencies. The attention computation is defined as:
\vspace{-2ex}
\begin{equation}
\mathrm{Att}(Q,K,V) = \mathrm{softmax}\left(\frac{QK^\top}{\sqrt{d}}\right)\cdot V
\label{eq:attention}
\end{equation}
where $Q$ (query), $K$ (key), and $V$ (value) are projections of either latent tokens or conditioning embeddings. 
While standard attention enables global interaction across all tokens, this becomes problematic in the multi-reference setting: cross-token mixing can lead to interference between unrelated entities—whether in textual descriptions, visual references, or layout controls. 

\subsection{Overview of LAMIC}
As shown in Figure~\ref{fig:framework}, LAMIC enables layout-aware multi-entity generation through the following three stages:
1) \underline{\textbf{Structured Input Definition}}, where each reference is organized into a visual-textual-spatial (VTS) triplet, complemented by cross-entity interaction (CEI) instructions and uncontrolled regions;
2) \underline{\textbf{Unified Token Representation}}, where all components—VTS triplets, CEI, and uncontrolled regions—are encoded into a unified token sequence for joint representation in MMDiT;
3) \underline{\textbf{Multi-VTS Guided Generation}}, where image synthesis is guided by all VTS tokens. Two attention mechanisms are introduced to support this stage: \textbf{Group Isolation Attention (GIA)} restricts cross-group interaction among textual, spatial, and visual tokens to prevent semantic entanglement; \textbf{Region-Modulated Attention (RMA)} defers inter-region fusion and CEI injection to enhance layout controllability and avoid early-stage semantic leakage.

\subsection{Structured Input Definition} 
We structure each reference as a visual-textual-spatial (VTS) triplet group $G_i=(V_i, T_i, S_i)$, where $V_i$ denotes the visual reference (image), $T_i$ represents the textual condition—referred to as the self-attribute description (SAD), and $S_i$ specifies the target spatial region (e.g., bounding box or mask). Each SAD consists of an \textbf{identifier} describing the entity (e.g., ``a dragon'', ``a car''), and a \textbf{description} specifying appearance behavior (e.g., ``keep the same appearance'', ``change the pose'').
We further introduce a \textbf{Cross-Entity Interaction (CEI)} instruction $C$, which governs spatial or semantic relationships between entities (e.g., ``A rides B''), and an \textbf{uncontrolled region} $U$, covering areas not assigned to any specific entity.
Together, these components—$\{(V_i, T_i, S_i)\}_{i=1}^N$, $C$, and $U$—form a structured input that aligns visual, textual, and spatial guidance, and $N$ denotes the number of references. This design supports multi-entity composition without relying on external reasoning modules such as MLLMs~\cite{yang2024mastering-RPG}.

\subsection{Unified Token Representation}
We encode all components—$\{(V_i, T_i, S_i)\}_{i=1}^N$, $C$, and $U$—into a unified token sequence. Specifically, we use the pretrained VAE or AE from MMDiT to convert each $V_i$ into latent tokens $L_i \in \mathbb{R}^{B \times (H_i \times W_i / 4) \times 4C}$, where $B$, $C$, $H_i$, and $W_i$ denote the batch size, channel count, and spatial dimensions. Textual inputs ($T_i$, $C$) are embedded via pretrained T5~\cite{LLMs-T5} or CLIP~\cite{CLIP}, and projected to match the latent token space. Spatial regions $S_i$ are downsampled (typically $\times8$) and reshaped to match the image token format. We then concatenate all tokens along the sequence dimension and record their positions for subsequent attention masking.

\subsection{Multi-VTS Guided Generation}
\begin{figure}[t]
    \centering
    \begin{subfigure}[b]{0.23\textwidth}
        \includegraphics[width=\linewidth]{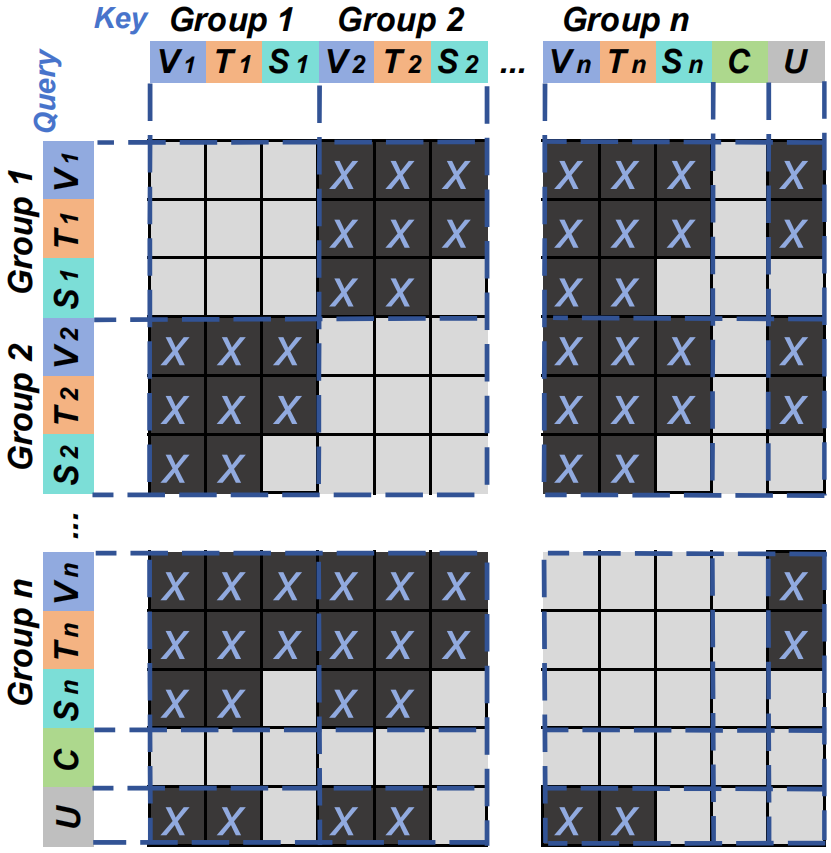}
        \caption{Group Isolation Attention}
        \label{subfig:group_isolation_attention}
    \end{subfigure}
    \begin{subfigure}[b]{0.23\textwidth}
        \includegraphics[width=\linewidth]{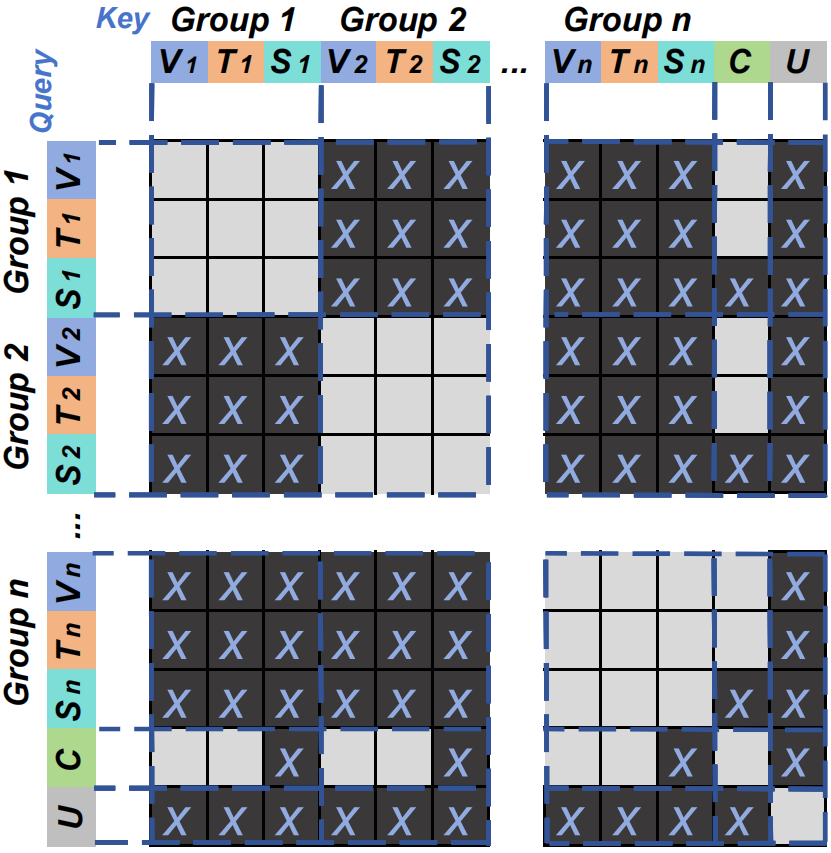}
        \caption{Region-Modulated Attention}
        \label{subfig:region_modulated_attention}
    \end{subfigure}
    \caption{Our proposed attention mechanisms.}
    \label{fig:attention_mechanisms}
\end{figure}

We design two attention mechanisms to support layout-aware generation guided by multi-VTS tokens: Group Isolation Attention (GIA) and Region-Modulated Attention (RMA), as illustrated in Figure~\ref{fig:attention_mechanisms}.

\paragraph{Group Isolation Attention (GIA).}
GIA suppresses interference across VTS groups by restricting attention computations within each group:
\begin{equation}
\label{GIA_G_G}
GIA(Q_{G_i},K_{G_j},V_{G_j})=
\begin{cases}
\mathrm{Att}(\cdot), & i = j \\
0, & i \ne j
\end{cases}
\end{equation}
where $i, j \in \{1, \dots, N\}$. For brevity, in multi-case equations, we use $\mathrm{Att}(\cdot)$ to denote the attention operation with the same $(Q, K, V)$ inputs as those on the left-hand side of the equation.

To ensure structural coherence, we retain unrestricted attention between spatial regions on the condition of Eq.~(\ref{GIA_G_G}):
\begin{equation}
GIA(Q_{S_i},K_{S_j},V_{S_j})=\mathrm{Att}(Q_{S_i},K_{S_j},V_{S_j}),~\forall i,j
\end{equation}
We also define cross-group interactions with CEI ($C$) and uncontrolled region ($U$): $C$ is treated as a global prompt and interacts fully with all groups, while $U$ follows the spatial attention pattern:
\begin{subequations}
\begin{align}
GIA(Q_{G_i}, K_{C}, V_{C}) &= \mathrm{Att}(Q_{G_i},K_{C},V_{C}) \\
GIA(Q_{C}, K_{G_i}, V_{G_i}) &= \mathrm{Att}(Q_{C},K_{G_i},V_{G_i})
\end{align}
\begin{align}
GIA(Q_{G_i^y},K_U,V_U) &= 
\begin{cases}
\mathrm{Att}(\cdot), & y=S \\
0, & y \ne S
\end{cases} \\
GIA(Q_U,K_{G_i^y},V_{G_i^y}) &= 
\begin{cases}
\mathrm{Att}(\cdot), & y=S \\
0, & y \ne S
\end{cases}
\end{align}
\end{subequations}
where $y \in \{V,T,S\}$, $\forall i \in N$. 

\paragraph{Region-Modulated Attention (RMA).}
To promote precise spatial control and prevent early-stage semantic leakage, based on GIA, RMA further limits the inter-region cross-attention and CEI injection in the early denoising step:
\begin{equation}
RMA(Q_{S_i},K_{S_j},V_{S_j})=
\begin{cases}
\mathrm{Att}(\cdot), & i = j \\
0, & i \ne j
\end{cases}
\end{equation}
and attention with $U$ and $C$ is disabled:
\begin{align}
RMA(Q_{S_i},K_U,V_U) &= RMA(Q_U,K_{S_i},V_{S_i}) = 0 \\
RMA(Q_{S_i},K_C,V_C) &= RMA(Q_C,K_{S_i},V_{S_i}) = 0 \\
RMA(Q_U,K_C,V_C) &= RMA(Q_C,K_U,V_U) = 0
\end{align}

In practice, we implement these rules via attention masks for efficiency. The total denoising process is divided into two sub-stages: RMA is applied during the first stage, covering a predefined ratio of the total steps, followed by GIA in the remaining steps.

\begin{figure*}[t!]
    \centering
    \includegraphics[width=\linewidth]{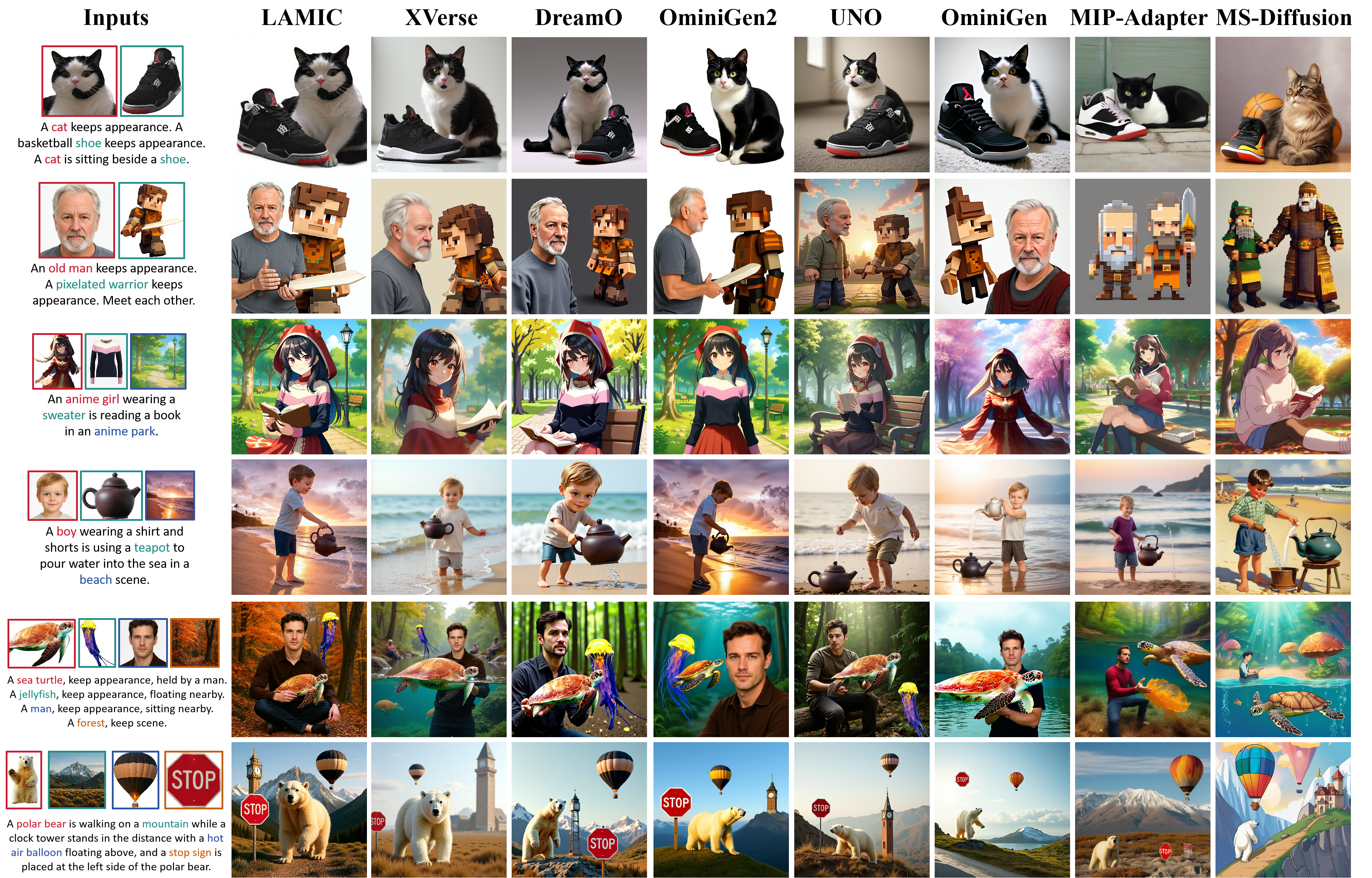}
    \caption{Visual comparison of different methods under different multi-reference images.}
    \label{fig:visual_comparison}
\end{figure*}

\section{Experiments}
\subsection{Experimental Setting}
\paragraph{Implementation Details.}
We implement LAMIC based on the open-source single-image reference MMDiT-based model (Flux.1 Kontext-dev). The inference process is configured with 20 denoising steps, a guidance scale of 2.5, and a first-stage step ratio of 0.05. To reduce memory consumption, both the Transformer and T5 modules are quantized to INT8 during inference. All experiments are conducted on a machine equipped with a single NVIDIA RTX 4090 and dual NVIDIA A6000 GPUs.

\paragraph{Benchmark Dataset.}
As a benchmark for multi-image composition, XVerseBench~\cite{chen2025xverse} originally includes 74 objects, 20 human faces, and 45 animals. However, we observed its limitations in subject diversity and visual quality. To address this, we augmented the dataset with 20 additional scenes, 17 clothing items, and 1 object sourced from DreamBench++~\cite{peng2024dreambench}, MS-Bench~\cite{wang2025msdiffusion}, and GPT-4o generations. Moreover, due to the low resolution and noise present in some original samples, we regenerated 20 high-resolution human faces and replaced several degraded images using high-quality generation tools. Beyond improving image quality and type diversity, we constructed structured multi-image inputs with associated bounding boxes for each subject, enabling precise layout-aware generation and evaluation. Specifically, we created 60 inputs with two reference images, 40  inputs with three reference images, and 20 inputs with four reference images. 


\paragraph{Evaluation Metrics.}
Following prior work~\cite{chen2025xverse}, we adopt several established metrics to assess generation quality. To better evaluate background consistency and layout controllability, we further propose three novel metrics: \textbf{BG-S}, \textbf{IN-R}, and \textbf{FI-R}, which provide finer-grained analysis in multi-reference, layout-aware synthesis settings.
\begin{itemize}
    \item \textbf{DPG Score}~\cite{metrics-DPGscore-hu2024ella}, measuring the text consistency editing ability of the model;
    \item \textbf{Face ID Similarity (ID-S)}~\cite{metrics-FaceIDsim-deng2019arcface}, evaluating human identity preservation;
    \item \textbf{DINOv2 Similarity (IP-S)}~\cite{metrics-DINOsim-oquab2023dinov2}, capturing object appearance consistency;
    \item \textbf{Aesthetic Score (AES)}~\cite{metrics-AESscore-discus0434_2024}, judging overall aesthetic appeal;
    \item \textbf{Background Similarity (BG-S)}, a weighted combination of DINOv2, CLIP~\cite{CLIP}, SSIM, and color histogram (CH):
\end{itemize}

\vspace{-15pt}
\begin{equation}
\scalebox{0.96}{$
\textit{BG-S} = 0.4 \cdot \textit{DINO}
    + 0.25 \cdot \textit{CLIP}
    + 0.2 \cdot \textit{SSIM}
    + 0.15 \cdot \textit{CH}
$}
\vspace{-2pt}
\end{equation}
We report the unweighted average of the above five metrics as \textbf{AVG} for overall generation quality.

For layout evaluation, we employ the Grounded SAM 2 pipeline, where Florence‑2~\cite{foundation_models-Xiao2024Florence2} handles object detection and grounding, generating bounding boxes for target entities, which are then processed by SAM‑2 to produce precise segmentation masks \(M_{\text{gen}}\), which are compared to the ground-truth target region mask \(M_{\text{trg}}\).

Specifically, \textbf{IN-R (Inclusion Ratio)} measures how much of the generated entity lies within the target region, while \textbf{FI-R (Fill Ratio)} evaluates how well the target region is covered by the generated entity. These two ratios jointly reflect the precision and completeness of layout control. In order to unify the value range with other metrics, we multiplied both ratios by 100.
\vspace{-1pt}
\begin{equation}
\textit{IN-R} = \frac{\sum (M_{\text{gen}} \cap M_{\text{trg}})}{\sum M_{\text{gen}}} \times 100
\end{equation}
\begin{equation}
\textit{FI-R} = \frac{\sum (M_{\text{gen}} \cap M_{\text{trg}})}{\sum M_{\text{trg}}} \times 100
\end{equation}

To avoid artificially inflated layout scores caused by large targets (e.g., full-image regions), we discard samples where the target mask occupies more than 75\% of the image area.



\paragraph{Baseline Methods.}
We first evaluate the performance of our method on multi-image composition tasks, comparing it against several state-of-the-art multi-reference generation approaches, including MS-Diffusion~\cite{wang2025msdiffusion}, MIP-Adapter~\cite{huang2025resolving_MIP_Adapter}, OmniGen~\cite{xiao2024omnigen}, UNO~\cite{wu2025less_UNO}, OmniGen2~\cite{wu2025omnigen2}, DreamO~\cite{mou2025dreamo}, and XVerse~\cite{chen2025xverse}. These models are evaluated across various subject composition and background integration scenarios under three settings. After that, we further evaluate the layout control performance of the above methods.

\begin{table*}[t!]
\centering
\caption{Quantitative results of multi-image combination. \textbf{Bold} indicates the best result, \underline{single underline} indicates the second-best, and \underline{\underline{double underline}} indicates the third-best.}
\label{tab:comparison}
\small
\setlength{\tabcolsep}{1.5pt}
\renewcommand{\arraystretch}{1.1}
\resizebox{\linewidth}{!}{
\begin{tabular}{l|cccccc|cccccc|cccccc}
\toprule
\multirow{2}{*}{\textbf{Method}} 
& \multicolumn{6}{c|}{\textbf{Two-Reference}} 
& \multicolumn{6}{c|}{\textbf{Three-Reference}} 
& \multicolumn{6}{c}{\textbf{Four-Reference}} \\
& DPG & ID-S & IP-S & BG-S & AES & AVG
& DPG & ID-S & IP-S & BG-S & AES & AVG
& DPG & ID-S & IP-S & BG-S & AES & AVG \\
\midrule
MS-Diffusion 
&75.01&12.70&47.05&71.48&52.88&51.82 
&86.67&3.46&43.13&72.49&57.85&52.72
&75.24&3.78&41.25&73.24&55.41&49.78\\
MIP-Adapter
&82.64&22.28&59.59&\underline{\underline{75.15}}&55.77&59.09
&84.97&20.40&61.58&78.93&\underline{60.61}&61.30
&83.49&13.31&57.73&77.82&\underline{\underline{58.60}}&58.19\\
OmniGen     
&82.06&\underline{69.32}&66.63&73.25&\underline{\underline{56.54}}&69.56
&72.79&61.24&64.52&78.30&\underline{\underline{59.60}}&67.29
&74.60&54.88&61.60&77.21&\underline{59.39}&65.54\\
UNO    
&\textbf{89.42}&38.71&\underline{72.92}&72.32&\textbf{59.52}&66.58
&\underline{91.00}&43.52&\underline{\underline{69.03}}&78.72&\textbf{62.45}&68.94
&\textbf{90.16}&39.98&\textbf{73.73}&78.34&\textbf{61.75}&68.79\\
OmniGen2    
&85.53&\underline{\underline{59.89}}&70.60&\underline{80.59}&54.61&\underline{70.24}
&81.00&55.53&65.88&\underline{83.58}&59.32&69.06
&81.09&50.66&62.90&\underline{81.56}&56.49&66.54\\
DreamO    
&\underline{88.54}&58.71&\textbf{73.45}&73.99&55.28&\underline{\underline{69.99}}
&\underline{\underline{90.60}}&\underline{63.74}&\underline{71.49}&\underline{\underline{80.73}}&57.01&\underline{\underline{72.71}}
&\underline{90.04}&\underline{\underline{57.29}}&\underline{\underline{70.19}}&\underline{\underline{80.08}}&55.95&\underline{70.71}\\
XVerse 
&\underline{\underline{87.90}}&56.49&70.72&74.62&\underline{59.38}&69.82 
&90.23&\underline{\underline{63.70}}&\textbf{73.19}&79.53&59.06&\underline{73.14} 
&\underline{\underline{84.00}}&\underline{61.84}&\underline{70.69}&78.72&56.71&\underline{\underline{70.39}}\\
\textbf{LAMIC(Ours)} 
& 85.61 & \textbf{78.04} & \underline{\underline{72.33}}&\textbf{83.14}&53.59&\textbf{74.54} 
&\textbf{91.95}&\textbf{65.63}&67.54&\textbf{86.06}&59.24&\textbf{73.92}
&\textbf{90.16}&\textbf{70.25}&66.67&\textbf{87.02}&58.10&\textbf{74.44}\\
\bottomrule
\end{tabular}
}
\end{table*}

\subsection{Multi-Image Composition Performance}
\label{exp:multi-image composition}
\paragraph{Quantitative Comparison.}
Table~\ref{tab:comparison} quantitatively demonstrates that LAMIC consistently achieves the best overall performance across all settings. Notably, it obtains the highest ID-S, BG-S, and AVG scores in each reference configuration, indicating strong identity and background preservation, as well as balanced generation quality. In the three- and four-reference settings, LAMIC also achieves the highest DPG score, demonstrating its superior editing capability and prompt consistency.
\textit{\underline{\textbf{(1) Specifically, in the two-reference setting}}}, LAMIC achieves an ID-S of 78.04, surpassing the second-best OmniGen by nearly 9 points; a BG-S of 83.14, exceeding the second-best OmniGen2 by 2.55; and an AVG of 74.54, outperforming the runner-up OmniGen2 by 4.3. Furthermore, the IP-S of 72.33 is only 1.12 below the best-performing model, demonstrating excellent object preservation.
\textit{\underline{\textbf{(2) In the three-reference setting}}}, LAMIC achieves a DPG of 91.95, an ID-S of 65.63, and an AVG of 73.92, all outperforming the respective second-best models by approximately 1 point. Notably, it attains a BG-S of 86.06, which is 2.5 points higher than the second-best OmniGen2, further validating LAMIC’s superior capability in background consistency.
\textit{\underline{\textbf{(3) In the more challenging four-reference setting}}}, LAMIC still leads with a DPG of 90.16 (tied with UNO), an ID-S of 70.25 (exceeding the second-best XVerse by 8.41), a BG-S of 87.02, and an AVG of 74.44, surpassing all competitors by a large margin, indicating LAMIC maintains strong performance as the reference number increases.

These improvements are achieved without any fine-tuning or model re-training, highlighting the zero-shot generalization capability of our method. It is also worth noting that UNO consistently achieves the best AES, reflecting its strong aesthetic appeal, while OmniGen2 performs second-best on BG-S, demonstrating good background synthesis ability. DreamO and XVerse also achieve solid overall performance, closely following LAMIC in terms of AVG.

\paragraph{Qualitative Comparison.}
Figure~\ref{fig:visual_comparison} presents qualitative comparisons under diverse multi-reference scenarios. LAMIC excels in preserving subject identity and structural fidelity, generating visually coherent and high-quality results. For example, in the ``old man-pixelated warrior'' composition (Row 2), LAMIC successfully maintains the subject’s stylized structure and realistic blending, while other methods exhibit over-smoothing or distortions. In the ``sea turtle–jellyfish–man–forest'' composition (Row 5), LAMIC respects spatial arrangements and visual semantics, accurately merging all referenced elements, whereas most baselines suffer from object mismatching or semantic drifting.

XVerse consistently delivers visually pleasing generations with relatively high ID-S, particularly in cases with prominent human references (Rows 2 and 4), but tends to oversimplify background compositions. DreamO achieves smoother image transitions and realistic style rendering, as observed in Row 3 (``anime girl'' etc.) and Row 4 (``teapot and beach'' etc.), but occasionally struggles with precise identity preservation and text following, especially in more complex scenes. In contrast, methods like MIP-Adapter and MS-Diffusion exhibit limitations in balancing layout, identity, and appearance, often leading to incomplete or mismatched object integration.

Overall, both quantitative and qualitative results validate that LAMIC achieves superior multi-image composition, maintaining generation quality and consistency. More examples are illustrated in \textit{Appendix A}.

\begin{table*}[t!]
\centering
\caption{Ablation results for different attentions and ratios of first-stage steps under layout-aware multi-image composition.}
\label{tab:lamic_ablation}
\small
\setlength{\tabcolsep}{1.5pt}
\renewcommand{\arraystretch}{1}
\resizebox{\linewidth}{!}{
\begin{tabular}{cc|ccccccc|ccccccc|ccccccc}
\toprule
\multicolumn{2}{c|}{\multirow{2}{*}{\textbf{Settings}}}
& \multicolumn{7}{c|}{\textbf{Two-Reference}} 
& \multicolumn{7}{c|}{\textbf{Three-Reference}} 
& \multicolumn{7}{c}{\textbf{Four-Reference}} \\
& & DPG & ID-S & IP-S & BG-S & AES & IN-R & FI-R 
  & DPG & ID-S & IP-S & BG-S & AES & IN-R & FI-R 
  & DPG & ID-S & IP-S & BG-S & AES & IN-R & FI-R \\
\midrule
\multirow{4}{*}{\rotatebox[origin=c]{90}{\textbf{LAMIC}}}
& \multicolumn{1}{|c|}{0.05} 
&85.61&78.04&72.33&83.14&\underline{53.59}&92.39&32.75
&\textbf{91.95}&65.63&\textbf{67.54}&86.06&\underline{59.24}&91.90&24.26
&90.16&\textbf{70.25}&\textbf{66.67}&\textbf{87.02}&58.10&89.81&\textbf{20.81}
\\
& \multicolumn{1}{|c|}{0.10} 
&\textbf{86.76}&79.73&\textbf{72.99}&84.74&52.73&94.34&31.98
&89.56&\underline{66.37}&66.76&86.33&\underline{59.24}&93.92&26.31
&\textbf{92.71}&\underline{66.85}&\underline{65.59}&86.14&58.36&89.27&20.23
\\
& \multicolumn{1}{|c|}{0.15} 
&\underline{86.26}&\underline{81.21}&72.90&\textbf{85.17}&52.39&\textbf{95.34}&\underline{32.94}
&88.84&\textbf{66.97}&\underline{67.23}&\underline{86.52}&59.09&\textbf{94.67}&\underline{27.33}
&89.50&66.81&65.57&86.10&58.17&\underline{90.25}&20.54
\\
& \multicolumn{1}{|c|}{0.20} 
&85.47&\textbf{81.37}&\underline{72.93}&84.48&51.90&\underline{95.26}&\textbf{33.43}
&86.48&65.53&67.15&\textbf{86.58}&59.17&\underline{94.65}&\textbf{27.49}
&89.14&66.80&\underline{65.59}&86.13&\underline{58.41}&\textbf{90.72}&\underline{20.71}
\\
\midrule
\multicolumn{2}{c|}{w/o RMA} 
&84.18&{67.96}&{67.25}&82.95&\textbf{54.91}&{81.77}&{28.93}
&\underline{90.92}&{64.58}&{64.51}&{85.79}&\textbf{59.31}&{87.45}&{23.81}
&\underline{91.25}&{65.81}&{64.84}&\underline{86.30}&\textbf{58.83}&{88.88}&{19.39}
\\
\multicolumn{2}{c|}{w/o GIA} 
&86.20&39.15&61.66&79.95&53.30&66.12&24.47
&87.30&42.19&55.60&84.16&57.63&69.37&20.70
&84.74&32.07&51.62&85.08&54.85&66.95&16.57
\\
\bottomrule
\end{tabular}
}
\end{table*}

\begin{table}[t!]
\centering
\caption{Comparative results of our LAMIC and other methods under layout-aware multi-image composition.}
\label{tab:comparision_layout_control}
\small
\setlength{\tabcolsep}{2pt}
\renewcommand{\arraystretch}{1.2}
\begin{tabular}{l|cc|cc|cc}
\toprule
\multirow{2}{*}{\textbf{Method}} 
& \multicolumn{2}{c|}{\textbf{Two-Ref}} 
& \multicolumn{2}{c|}{\textbf{Three-Ref}} 
& \multicolumn{2}{c}{\textbf{Four-Ref}} \\
& IN-R & FI-R 
& IN-R & FI-R 
& IN-R & FI-R \\
\midrule
MS-Diffusion 
&56.83&23.17
&72.84&19.06
&58.55&20.32\\
MIP-Adapter
&59.25&21.73
&70.43&20.34
&\underline{\underline{63.16}}&19.40\\
OmniGen
&58.60&16.40
&66.43&20.87
&58.99&14.96\\
UNO
&63.46&15.74
&70.37&18.54
&\underline{70.13}&17.85\\
OmniGen2
&\underline{67.49}&\underline{27.87}
&70.72&\textbf{27.27}
&58.50&\textbf{22.30}\\
DreamO
&\underline{69.25}&\underline{\underline{24.37}}
&\underline{\underline{73.84}}&\underline{\underline{23.57}}
&72.95&\textbf{23.71}\\
XVerse 
&62.65&20.30
&\underline{75.42}&22.83
&62.24&19.62\\
\textbf{LAMIC(Ours)} 
& \textbf{92.39} & \textbf{32.75}
& \textbf{91.90} & \underline{24.26}
& \textbf{89.81} & \underline{\underline{20.81}} \\
\bottomrule
\end{tabular}
\end{table}

\subsection{Layout-Controlled Multi-Image Composition Performance}
\label{sec:layout-control}
Although MS-Diffusion is currently the only baseline that explicitly supports layout-aware multi-image composition, we still compare our method against all aforementioned approaches to provide a comprehensive evaluation of our proposed layout control metrics.

The results are presented in Table~\ref{tab:comparision_layout_control}. LAMIC achieves an IN-R of approximately 90 across all settings, significantly outperforming all other methods. This is expected, as most baselines lack explicit layout control capabilities. While MS-Diffusion claims to support layout-aware generation, its performance on both IN-R and FI-R is relatively subpar. We attribute this to the nature of our proposed IN-R and FI-R metrics, which evaluate the consistency of entity placement and spatial accuracy. These metrics rely on a model’s ability to preserve entities and maintain compositional alignment—areas where MS-Diffusion underperforms (Sec.~\ref{exp:multi-image composition}), thus leading to lower scores.

It is also worth noting that although LAMIC consistently achieves the best FI-R among all methods, the margin over layout-unaware baselines is not large. This suggests that while LAMIC demonstrates superior layout control, there is still considerable room for improvement in fully capturing and preserving target spatial configurations. 

\subsection{Ablation Study}

In our ablation study, we first analyze the impact of removing each proposed module, and then investigate the effect of varying the ratio of first-stage steps on generation quality.
\begin{figure}[ht]
    \centering
    \includegraphics[width=\linewidth]{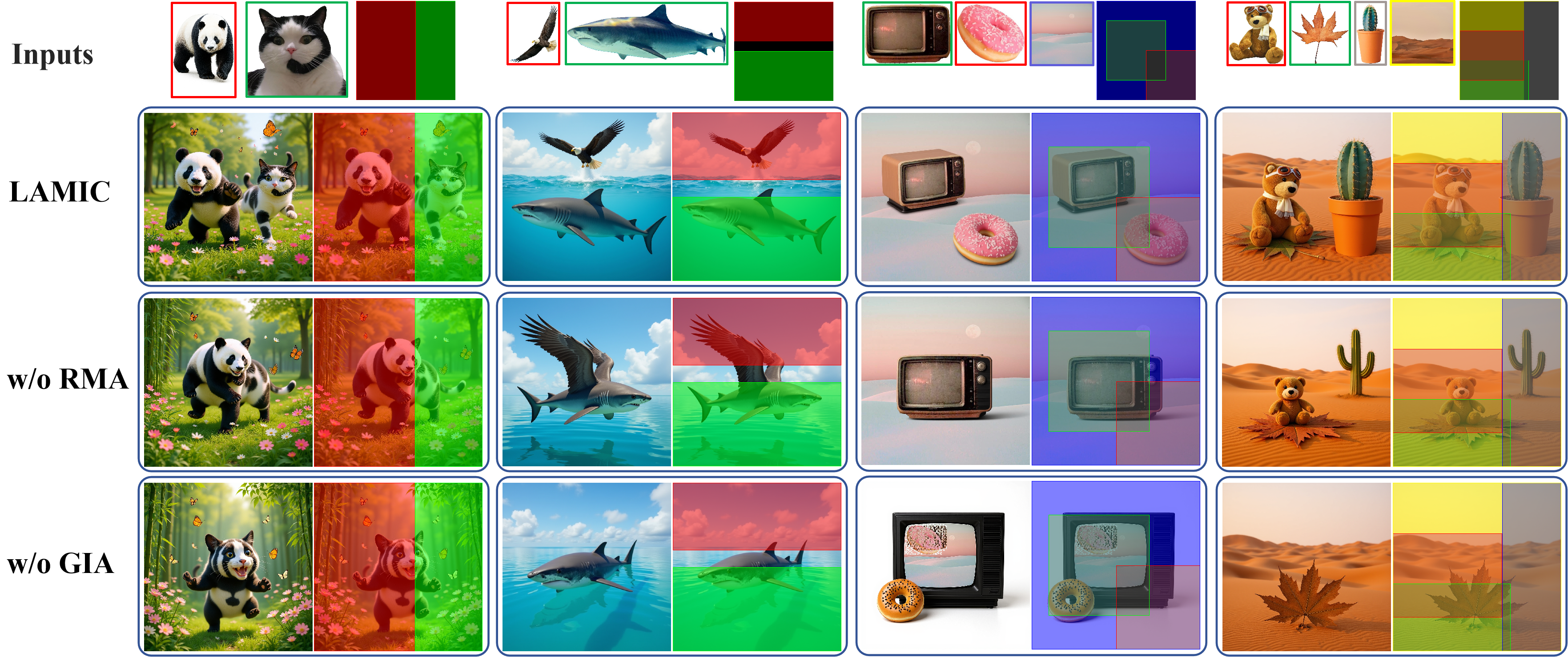}
    \caption{Visual comparison of different settings of LAMIC under layout-aware multi-image composition.}
    \label{fig:visualize_module_ablation}
\end{figure}

\paragraph{Impact of Proposed Modules.}
Table~\ref{tab:lamic_ablation} quantitatively demonstrates the impact of individually removing the proposed RMA and GIA components. The full model configuration (LAMIC) achieves the best overall performance. Using RMA results in a slight drop in aesthetic quality, which is reasonable given the substantial improvement in layout control—IN-R increases by up to 10.62 points in the two-reference setting and 7.22 points in the three-reference setting compared to the version without RMA. In contrast, removing GIA causes a significant degradation across nearly all metrics, highlighting its critical role in ensuring high-quality multi-image composition.

Figure~\ref{fig:visualize_module_ablation} presents qualitative comparisons across different settings. It clearly shows that removing RMA weakens layout control performance (e.g., in the ``TV-donut'' and ``bear-maple leaf-cactus-desert'' cases), and leads to partial fusion or entanglement of entities within target regions (e.g., in ``panda-cat'' and ``eagle-shark''). The degradation becomes more pronounced when GIA is removed: layout control capabilities nearly vanish (consistent with the scores in Table~\ref{tab:lamic_ablation}), which are comparable to layout-unaware baselines Table~\ref{tab:comparision_layout_control}, and multiple reference entities collapse into a single blended form in most situations or just keep a single entity.

 \paragraph{Impact of First Stage Steps.}
 \begin{figure}[ht!]
     \centering
     \includegraphics[width=\linewidth]{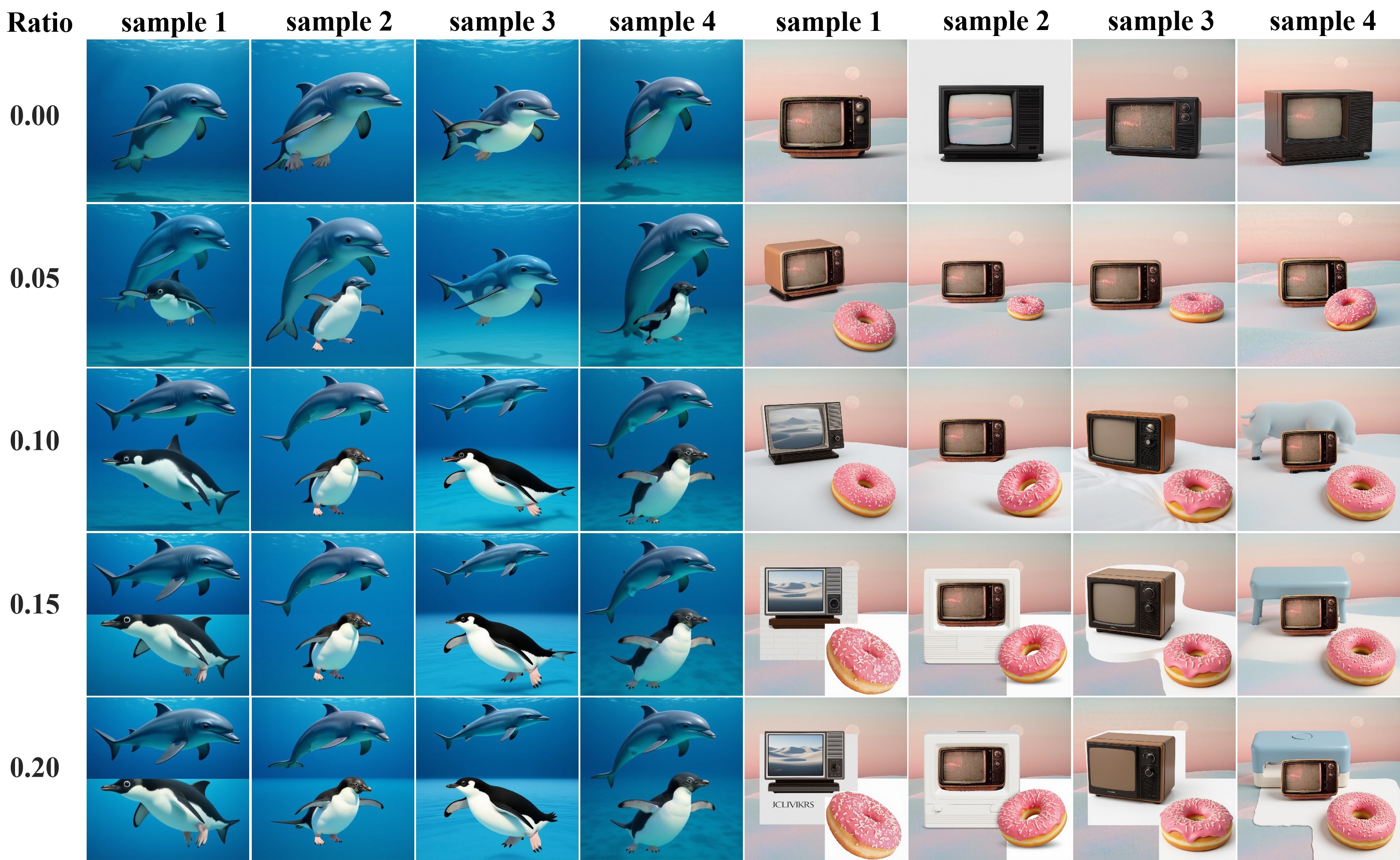}
     \caption{Visual comparison of different first-stage step ratios in LAMIC under layout-aware multi-image composition. For the left group, the target regions are the upper and lower halves (0.5 each); for the right group, the target regions correspond to those shown in Figure~\ref{fig:visualize_module_ablation}.}
     \label{fig:visual_ablation_ratio}
 \end{figure}
 
Table~\ref{tab:lamic_ablation} also reveals the impact of varying the ratio of first-stage steps. As expected, increasing this ratio generally improves layout control: in both two-, three-, and four-reference settings, a ratio of 0.15 or 0.20 consistently achieves the best or second-best scores in IN-R and FI-R. An exception occurs in the four-reference case, where a lower ratio of 0.05 yields the highest FI-R.

However, this improvement in layout control comes at the cost of other aspects of generation quality. As the ratio increases, the AES tends to decline, with the best scores observed at lower ratios (0.05 or 0.00, the latter corresponding to the w/o RMA case). A similar trend is observed for DPG, suggesting that excessive first-stage emphasis may impair prompt-following fidelity and regional continuity.

To visualize these effects, we present two groups of image samples in Figure~\ref{fig:visual_ablation_ratio}, where each sample in a single group shares identical inputs but uses different random seeds. These examples highlight the trade-off between layout precision and global coherence when adjusting the first-stage ratio. We further observe that when the ratio reaches 0.10 or higher, distinct boundaries begin to appear between adjacent regions, and visual consistency within individual regions degrades noticeably. In contrast, setting the ratio to 0.05 preserves better global coherence and intra-region consistency. Although the lower ratio may occasionally result in attribute blending between closely placed entities, its quantitative scores remain comparable and consistently yield superior visual quality in practice. Therefore, we fix the first-stage ratio to 0.05 in the main experiments.

\begin{figure*}[t!]
    \centering
    \includegraphics[width=\linewidth]{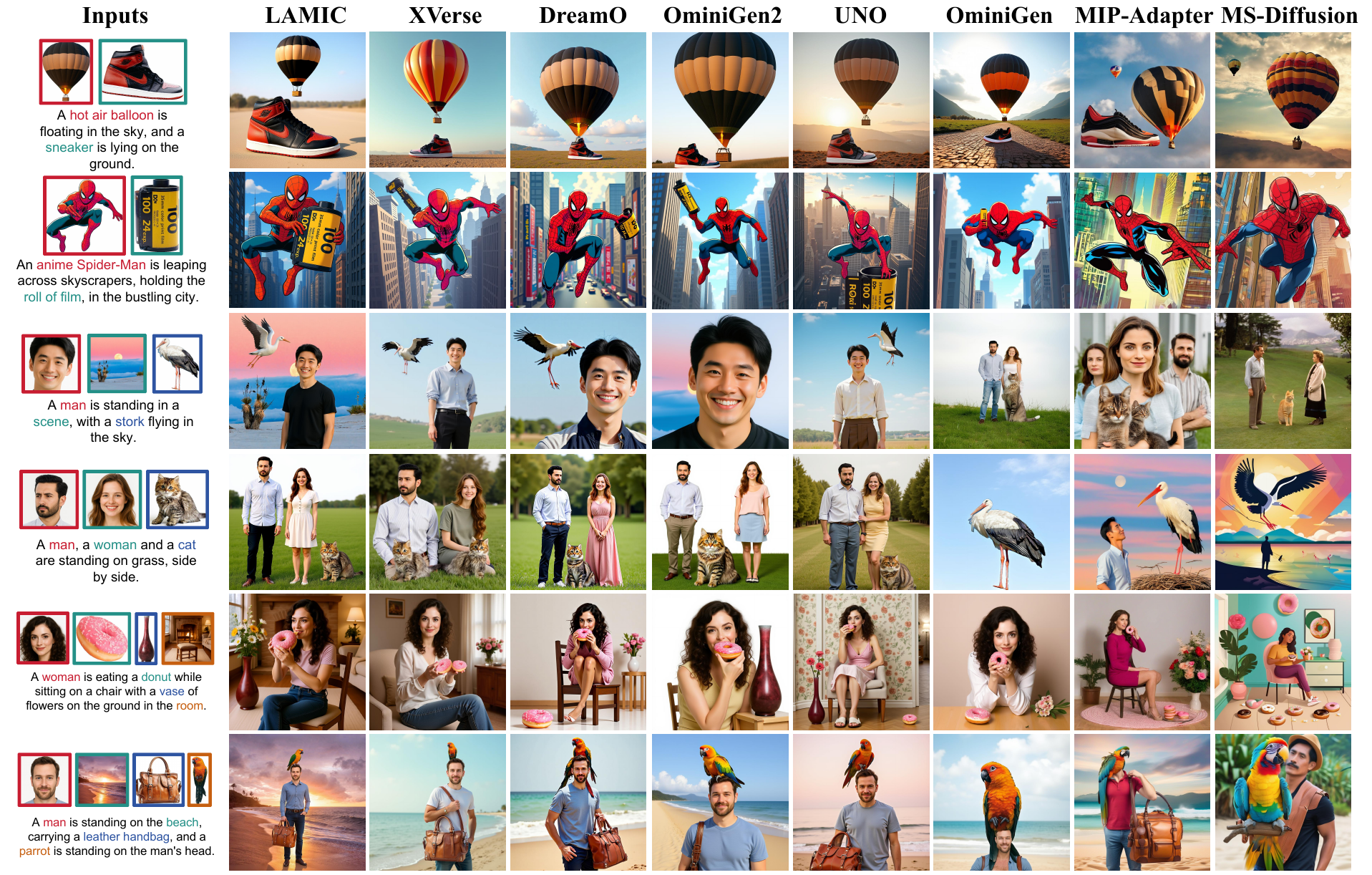}
    \caption{Supplementary visual comparisons between LAMIC and baseline models under diverse multi-reference compositions. LAMIC maintains better identity fidelity and background preservation across complex scenes.
}
    \label{fig:sup_visual_comparison}
\end{figure*}
\section{Conclusion}
In this work, we propose \textbf{LAMIC}, zero-shot framework for layout-aware multi-image composition. LAMIC is the first to extend the capabilities of consistent single-reference generative models to multi-reference generation and further introduces layout controllability—both without fine-tuning. To address semantic entanglement across references and enable precise region-wise control, we introduce two plug-and-play mechanisms: Group Isolation Attention (GIA) and Region-Modulated Attention (RMA). To more comprehensively quantify models' capabilities, we further introduce three metrics—IN-R, FI-R, and BG-S—the first two assess layout control, while the last measures background consistency.
Extensive experiments show that LAMIC achieves \textbf{state-of-the-art performance} across most of the major metrics. It consistently ranks first in ID-S, BG-S, IN-R, and AVG across 2-, 3-, and 4-reference settings, and leads in DPG under 3- and 4-reference scenarios. These results highlight its superior identity fidelity, spatial controllability, and prompt-following. In the future, we will focus on refined attention designs to reduce such confusion while preserving boundary smoothness. We also plan to explore prompt-to-reference binding for earlier Cross-Entity Interaction (CEI), enhancing entity interplay and language controllability in early stages. Exploring more effective methods to extend pre-trained single-reference foundation models to multi-reference settings is also a promising future direction.

\vspace{-2pt}
\begin{figure*}[t!]
    \centering
    \includegraphics[width=\linewidth]{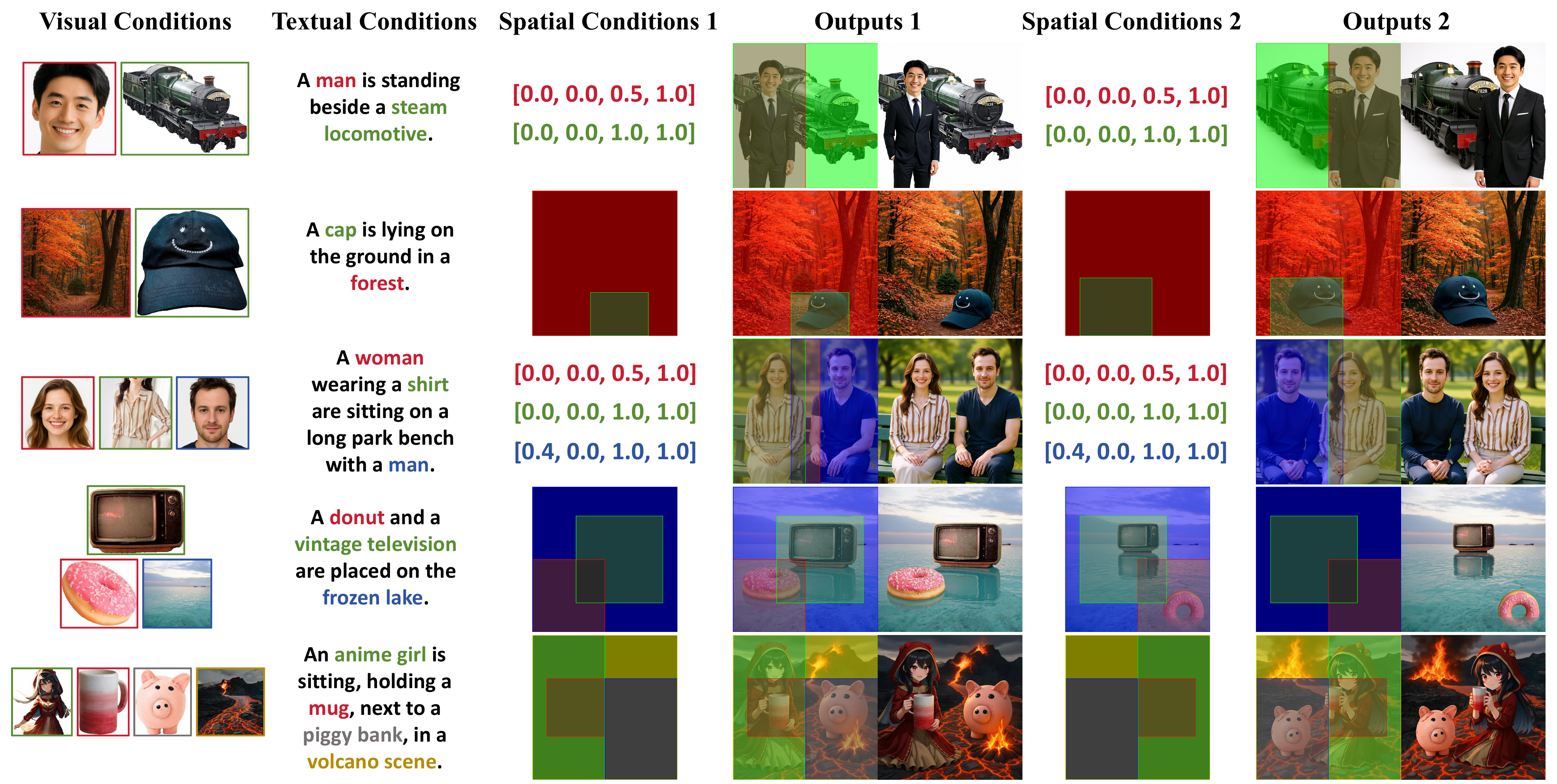}
    \caption{Visual illustration of LAMIC in layout-aware multi-image composition.}
    \label{fig:sup_visual_layout_control}
\end{figure*}

\section*{Appendix}
\subsection*{A. Supplementary Visual Comparisons in Multi-Image Composition}

Figure~\ref{fig:sup_visual_comparison} showcases additional visual comparisons across a wide range of multi-reference composition scenarios, further validating the performance of LAMIC in diverse conditions. Compared to existing methods, LAMIC consistently produces visually coherent results with accurate spatial placement and stronger identity preservation.

In the ``balloon-sneaker'' composition (Row 1), LAMIC generates a clean, layout-respecting integration of the two objects, while several baselines (e.g., MS-Diffusion, MIP-Adapter) show background clutter or degraded blending. In the ``Spider-Man–yellow sign–city'' setting (Row 2), only LAMIC and UNO maintains the stylized character structure and correctly embeds both referenced elements into the urban context; other models often distort spatial proportions or overly simplify comic textures.

In more entity-heavy scenes, such as the ``man–woman–cat'' (Row 4), LAMIC balances spatial arrangement and visual detail, preserving semantic separation among characters. In contrast, other methods tend to fuse identities or misplace key subjects. Similarly, in the “donut–vase–woman” indoor scene (Row 5), LAMIC successfully incorporates prompt following (e.g., vase position) while maintaining photorealistic composition, whereas most competitors produce inconsistent or missing characters or objects.

The ``man-scene-stork'' example (Row 3) and the ``man–leather handbag–parrot'' example (Row 6) highlight LAMIC’s robustness in fine-grained multi-entity scenarios. Most methods can generate consistent spatial stacking (e.g., the stork flying in the sky, and the parrot correctly placed on the head or shoulder), but only LAMIC avoids visual entanglement across distinct appearances, keeping natural synthesis and consistent entities.

Overall, these examples further demonstrate LAMIC’s advantage in consistent synthesis and prompt following across various multi-reference configurations.

\subsection*{B. Visual Illustration in Layout-Aware Multi-Image Composition}

Figure~\ref{fig:sup_visual_layout_control} provides a step-by-step illustration of LAMIC’s behavior under varying spatial conditions. Each row corresponds to a unique multi-image composition task with two sets of spatial configurations. The columns sequentially present visual references, textual prompts, spatial layout inputs, and resulting outputs from LAMIC.

In the ``man-steam locomotive'' example (Row 1), LAMIC preserves semantic coherence across both layouts. When the man's bounding box is moved from left to right (Spatial 1 vs. Spatial 2), the generated image correctly reflects this spatial instruction, while still maintaining his interaction with the locomotive. Notably, the man's gaze and pose adapt naturally to the layout change.

The ``cap-forest'' example (Row 2) demonstrates LAMIC's ability to handle scale and background harmonization. The spatial box constraining the cap shifts slightly, and the model adjusts the object placement while ensuring that both elements remain grounded in the forest scene with consistent color and shading.

In the ``woman-shirt-man'' case (Row 3), LAMIC aligns multiple entities to precise region constraints. The output respects relative positioning, ensuring that the woman and man are seated side by side in both layouts. The resulting composition maintains identity integrity and avoids character overlap, reflecting accurate triplet disentanglement.

Lastly, the ``donut-vintage television-frozen lake'' composition (Row 4) highlights LAMIC's capacity to resolve multiple prompts in a complex setting. The vintage television and donut are successfully grounded in distinct spatial regions across both configurations, and the frozen lake background is smoothly preserved. These results indicate that LAMIC can simultaneously track multiple bounding boxes and enforce spatial layout while retaining visual fidelity.

The ``anime girl–mug–piggy bank–volcano'' composition (Row 5) demonstrates LAMIC’s capability in handling more complex visual semantics under stylized and high-contrast backgrounds. Across both spatial configurations, the model accurately localizes each entity—positioning the girl, mug, and piggy bank into the prescribed regions—while preserving visual harmony with the intense volcanic environment. The generated entities maintain consistent appearance in interaction. Furthermore, the system successfully disentangles overlapping regions (e.g., mug and anime girl) and adapts the fiery scene lighting to all foreground elements, underscoring its robustness in both fine-grained identity control and background-aware rendering.

These visual illustrations confirm that LAMIC not only supports flexible layout input but also preserves compositional intent across different region constraints, validating its plug-and-play layout-aware generation capability.

\section*{License}
This preprint is made available under an arXiv.org non-exclusive license. The copyright remains with the authors.

{\small
\bibliographystyle{ieeenat_fullname}
\bibliography{references}
}

\end{document}